# Optimistic Variants of Single-Objective Bilevel Optimization for Evolutionary Algorithms

Anuraganand Sharma

*Abstract*—Single-objective bilevel optimization is a specialized form of constraint optimization problems where one of the constraints is the optimization problem itself. These problems are typically non-convex and strongly NP-Hard. Recently, there has been an increased interest from evolutionary computation community to model bilevel problems due to its applicability in the real world applications for decision-making problems. In this work, we are utilizing a partial nested evolutionary approach with local heuristic search to solve the benchmark problems and have outstanding results. This approach relies on the concept of intermarriage-crossover in search for feasible regions by exploiting information from the constraints. We are also proposing new variants to the commonly used convergence approaches, i.e., optimistic and pessimistic. The experimental results demonstrate the algorithm converges differently to known optimum solutions with the optimistic variants. Optimistic approach also outperforms pessimistic approach. Comparative statistical analysis of our approach with other recently published partial to complete evolutionary approaches demonstrates very competitive results.

## 1. INTRODUCTION

Optimization problems are very common in many fields such as engineering, operations research, economics, and games, to get the most favorable solution from a feasible set of solutions. Bilevel optimization problem is structured as a nested optimization problem in the form of constraint optimization problem (COP) where a solution must satisfy a set of constraints. These constraints can be either scalar/vector or static/dynamic functions [18]. In bilevel problems, one or more constraints are the COP themselves. This makes bilevel programming combinatorial and strongly NP-hard [8, 9]. Bilevel programming was motivated by [28] who used game theory to solve unbalanced economic market problem with multilevel optimization. Here, upper-level is called the *leader* and the lower-level is called the *follower*. The earliest formalization of bilevel problem using mathematical programming was published in [3] where bilevel problem was called *two-sided optimization* with *outside optimizer* and *inside optimizer* in a hierarchical setup.

Bilevel problem formulations employ either optimistic or pessimistic convergence approach where both have different selection criteria for selecting promising solutions; consequently, they result in different convergence [27, 29, 33, 34]. We have analyzed these convergence strategies in this paper and proposed new variants for better convergence for selected problems. Recently, there has been an increase in interest in solving bilevel optimization problem where many new classical and evolutionary approaches have been proposed [5, 7, 10, 22–27]. We concisely describe some commonly used methodologies for solving bilevel problems in the following section. Readers may find the detailed survey in [5, 27, 31].

A commonly used approach in mathematical programming is to convert bilevel optimization problem into a single level COP with Karush-Kuhn-Tucker (KKT) conditions, however, it may not necessarily be simple to handle especially when the upper-level's constraint functions are in an arbitrary linear form [2, 22]. Another commonly used approach is the penalty function methods where constraints are transformed into penalty terms, which in turn are used for reward and/or punishment for satisfying and/or violating the constraints, respectively [32]. However, its main shortcoming is that penalty factors that determine the severity of the punishment must be set by the user and their values are problem dependent [19]. Similar to penalty functions, a trust region method uses an initial approximation of a trust-region that expands (reward) if the approximation is good or contracts (punishment) otherwise. This can be called an iterative guided approach through trust-regions [13]. Another iterative reduction of approximation error is done with gradient descent approach [16]. The direction of descent leads towards optimization (least error) of the upper-level function while keeping the lower-level feasible. Gradient descent is commonly used in machine learning algorithms [1].

Finally, Evolutionary Algorithms (EAs) have also been used either in upper-level or both levels of bilevel problems [11, 14]. EAs are known to give reasonable solutions for NP-Hard optimization problems and they have been successfully applied to various forms of COPs [18]. Li et al. [11] have proposed a hierarchical particle swarm optimization for solving bilevel programming (HPSOBLP) that simulates the decision process of bilevel programming on both levels using Particle Swarm Optimization. However, this technique may have high time complexity because of the nested nature of the algorithm. Recently, some hybrid approaches of EAs with other mathematical techniques have been proposed in [26] and [10] where upper level is an EA and lower level is a local search. Sinha and Deb [23–27] have done an extensive work for evolutionary bilevel problems. They have proposed a Bilevel EA based on quadratic approximation (BLEAQ) that reduces the bilevel optimization problem to a single level optimization problem using quadratic functions [26]. They have also prepared a set of ten benchmark problems from the literature [26]. Kieffer et al. [10] have used Differential Evolution based Bayesian Optimization for Bilevel Problems (BOBP). BOBP has focused more on efficiency of the algorithm that also shows competitive results compared to BLEAQ.

Bilevel problem can have either single-objective optimization (SOO) or multi-objective optimization (MOO) for both the levels. We have focused on SOO for both the levels in this paper where a variation of EA has been applied to various convergence methodologies described for bilevel problems. For rest of the paper, we will refer single-objective bilevel optimization problem as bilevel problem only. We have enhanced Intelligent Constraint Handling Evolutionary Algorithm (ICHEA) [18, 19] that uses intermarriage-crossover twice in a generation; once each for both the levels of bilevel problems unlike BLEAQ, BLOP and HPSOBLP. ICHEA was designed to solve static and dynamic constraints effectively [18–21], however, it was never tested for a bilevel problem which is a special kind of COP. The enhanced ICHEA uses an evolutionary approach at the upper level and heuristic local search at the lower level with very promising test results. We have used ICHEA to analyze various kinds of convergence approaches on benchmark bilevel problems. The remainder of this paper is organized as follows: Section 2 describes the mathematical formulation of bilevel problems. Section 3 establishes the existing and our proposed variants for convergence techniques applicable for bilevel problems. Section 4 formalizes and describes the complete evolutionary approach to solve bilevel problems with ICHEA. Section 5 elaborates experimental results of our approach with three other recent partial and complete evolutionary approaches on benchmark problems and Section 6 concludes the paper by proposing future investigations.

## 2. Formulation of Bilevel Problems

Generally, a Bilevel problem is a two-level nested constraint optimization problem (COP). Hence, we initially define the formulation of COP. A COP is simply an optimization problem with a set of constraints. We have assumed that both levels are minimization problems for simplicity. Eq. (1) is minimization of COP's objective function $f(\vec{x})$ that has an $n$-dimensional input vector $\vec{x} = \{x_1, x_2, \dots x_n\}$ that is defined in a search space $S$.

$$\min f(\vec{x}) \tag{1}$$

More specifically, $\vec{x} \in \mathcal{F} \subseteq S$, where $\mathcal{F}$ being the feasible region on the search space $S \subseteq \mathbb{R}^n$. The domain of variables is defined by their lower bounds $l_i$ and upper bounds $u_i$:

$$l_i \leq x_i \leq u_i, \quad 1 \leq i \leq n \tag{2}$$

The feasible region $\mathcal{F}$ with bounds on each dimension is further restricted by a set of $m$ additional constraints that can be given in two relational forms – equality and inequality [6, 12, 17, 30].

$$g_i(\vec{x}) \geq 0 \qquad i = 1, \dots, k \tag{3}$$

$$h_j(\vec{x}) = 0 \quad j = k+1, \dots, m \tag{4}$$

The equality constraints $h_j(\vec{x})$ cannot be solved directly using EAs so it is converted into inequality constraints by introducing a positive tolerance value $\delta$.

$$g_j(\vec{x}) = \delta - |h_j(\vec{x})| \geq 0 \tag{5}$$

A set of individual feasible regions $\{\mathcal{F}_1, \mathcal{F}_2, \dots \mathcal{F}_m\}$ for each constraint can also be defined as:

$$\mathcal{F}_i = \{\vec{x} \in \mathcal{F} \mid g_i(\vec{x}) \geq 0, 1 \leq i \leq m, i \in Z\} \tag{6}$$

where $Z$ is the set of integers. Many EAs use a distance function as their fitness function to rank individuals. The distance function indicates how far a chromosome is from the feasible region [15]. This fitness function tries to bring the chromosomes closer to the feasible region using the following function for $\forall i : \{1 \leq i \leq m\}$:

$$fitness_i(\vec{x}) = \begin{cases} g_i(\vec{x}), & if \ g_i(\vec{x}) < 0 \\ 0, & if \ g_i(\vec{x}) \geq 0 \end{cases} \tag{7}$$

$$e = \sum_{i=1}^{m} |fitness_i(\vec{x})| \tag{8}$$

The fitness function $fitness_i$ in Eq. (7) is a measure of infeasibility of $\vec{x}$ from a feasible region $\mathcal{F}_i$. The error function $e$ is the summation of all the fitness functions as shown in Eq. (8). Minimizing the error value $e$ leads toward a constraint satisfaction problem's (CSP) solution where the objective function $f(\vec{x})$ is not needed. A solution to CSP is found when $e = 0$ or $\bigcap_{i=1}^{m} \mathcal{F}_i \neq \emptyset$. To get a COP solution, CSP solutions are further processed to get optimum value of $\vec{x}$ that optimizes the objective function $f(\vec{x})$.

Bilevel problem is simply a hierarchical set up of nested COP where the upper level is commonly known as the leader while the lower level is known as the follower. We have used the same variable names discussed above only with addition of subscripts $u$ and $v$ to indicate upper and lower level, respectively. The formulation is described below:

$\min F(\vec{x}_u, \vec{x}_l)$
Such that:
$\quad \underset{\vec{x}_l}{\text{argmin}}\, G_1(\vec{x}_u, \vec{x}_l)$
$\quad$ Such that:
$\qquad g_i(\vec{x}_u, \vec{x}_l) \geq 0 \quad i = 1, \dots, k_l$
$\qquad h_j(\vec{x}_u, \vec{x}_l) = 0 \quad j = k_l + 1, \dots, m_l$
$\quad G_i(\vec{x}_u, \vec{x}_l) \geq 0 \quad i = 2, \dots, k_u$
$\quad H_j(\vec{x}_u, \vec{x}_l) = 0 \quad j = k_u + 1, \dots, m_u$

where $F(\vec{x}_u, \vec{x}_l)$ is the upper level objective function with $m_u$ constraints that has one constraint in the form of lower level objective function given as $\underset{\vec{x}_l}{\text{argmin}}\, G_1(\vec{x}_u, \vec{x}_l)$. It is generally written as $\underset{\vec{x}_l}{\text{argmin}}\, f(\vec{x}_u, \vec{x}_l)$ to be consistent with the upper level objective function [5, 10, 27].

## 3. Variants of Convergence Techniques for Bilevel Problems

Two widely discussed optimization variants are the optimistic and the pessimistic models [27, 29, 33, 34]. We discuss these variants with our proposed variants that an EA can use in a given generation in the following section. These variants demonstrate variable extent of greediness in their choice for selecting a solution. The population converges differently with these strategies demonstrated in Section 5.

### 3.1. Optimistic w.r.t upper level ($O_F$)

In this greedy approach, the follower altruistically chooses a feasible solution that benefits the leader the most.

$$\Psi(\vec{x}_u) = \left\{ \vec{x}_l : \vec{x}_l \in \underset{\vec{x}_l}{\text{argmin}}\, f(\vec{x}_u, \vec{x}_l) : (\vec{x}_u, \vec{x}_l) \in \mathcal{F}_l \right\}$$

If $\Psi(\vec{x}_u)$ is not a singleton then:

$$\Psi^o(\vec{x}_u) = \left\{ \vec{x}_l \in \underset{\vec{x}_l}{\text{argmin}}\, F(\vec{x}_u, \vec{x}_l) : \vec{x}_l \in \Psi(\vec{x}_u) \right\}$$

Again, we may not get a singleton set but the "best" is picked at random. The overall optimistic model w.r.t upper level can be defined as:

$\min F(\vec{x}_u, \vec{x}_l)$
Such that:
$\quad \vec{x}_l = \Psi^o(\vec{x}_u)$
$\quad (\vec{x}_u, \vec{x}_l) \in \mathcal{F}_u$

To illustrate further, a numerical example is given in TABLE I, which shows sample fitness values for a given $\vec{x}_u$ and its corresponding parameter $\vec{x}_{l_i}$ where $i = 1 \dots 5$. The minimum lower level fitness is 100 and the corresponding minimum upper level fitness is 22. Hence, $O_F$ approach will result in the selection of $(\vec{x}_u, \vec{x}_{l_2})$.

TABLE I
Sample Data for Convergence Variants

| | | $f(\vec{x}_u, \vec{x}_{l_i})$ | $F(\vec{x}_u, \vec{x}_{l_i})$ | |
|---|---|---|---|---|
| $P_F \rightarrow$ | $\vec{x}_u, \vec{x}_{l_1}$ | 100 | 25 | |
| $O_F \rightarrow$ | $\vec{x}_u, \vec{x}_{l_2}$ | 100 | 22 | |
| | $\vec{x}_u, \vec{x}_{l_3}$ | 100 | 23 | |
| | $\vec{x}_u, \vec{x}_{l_4}$ | 112 | 11 | $\leftarrow EO_F$ |
| | $\vec{x}_u, \vec{x}_{l_5}$ | 115 | 50 | |

Conversely, the other variant for optimistic w.r.t lower level ($O_l$) can also be easily formulated where a solution chosen by the leader will benefit the follower the most.

### 3.2. Pessimistic w.r.t upper level ($P_F$)

On this convergence approach, the follower shows no cooperation to the leader and gives the worst feasible solution from the lower level.

$$\Psi(\vec{x}_u) = \left\{ \vec{x}_l : \vec{x}_l \in \underset{\vec{x}_l}{\operatorname{argmin}} f(\vec{x}_u, \vec{x}_l) : (\vec{x}_u, \vec{x}_l) \in \mathcal{F}_l \right\}$$

If $\Psi(\vec{x}_u)$ is not a singleton then:

$$\Psi^{\mathrm{p}}(\vec{x}_u) = \left\{ \vec{x}_l \in \underset{\vec{x}_l}{\operatorname{argmax}} F(\vec{x}_u, \vec{x}_l) : \vec{x}_l \in \Psi(\vec{x}_u) \right\}$$

The complete pessimistic $P_F$ model would be:

min $F(\vec{x}_u, \vec{x}_l)$
Such that:
$\quad \vec{x}_l = \Psi^{\mathrm{p}}(\vec{x}_u)$
$\quad (\vec{x}_u, \vec{x}_l) \in \mathcal{F}_u$

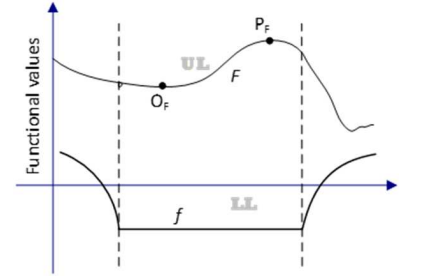

Fig. 1. A sketch to illustrate optimistic and pessimistic conditions w.r.t. upper-level

In TABLE I, the worst feasible solution for the upper level would be produced by parameters $\overrightarrow{x}_u, \vec{x}_{l_1}$. Fig.1 illustrates the difference between $O_F$ and $P_F$ when $\Psi(\vec{x}_u)$ is not a singleton.

Wiesemann et al. [33] has defined that the follower may choose to give any possible feasible solution not necessarily the best or the worst. In this case:

$$\Psi^{\mathrm{p}}(\vec{x}_u) = \forall \{\vec{x}_l \in \Psi(\vec{x}_u)\}$$

In this variant, a set of solutions for $P_F$ in TABLE I would be $(\vec{x}_u, \vec{x}_{l_i})$ with $i = 1 \dots 3$ where any one of the solution can be picked randomly.

### 3.3. Extreme optimistic w.r.t upper level ($EO_F$)

In this extremely greedy model, w.r.t upper level, the follower blindly chooses a partially feasible solution $(\vec{x}_u, \vec{x}_l^*)$ that benefits the leader the most, disregarding the other more promising solutions available for himself at a given generation. However, the chosen lower level solution should not be worse than the previously found lower level solution in anticipation that optimum solution may be found later on.

$$\Omega(\vec{x}_u) = \left\{ \vec{x}_l^* : f(\vec{x}_u, \vec{x}_l^*) \le f(\vec{x}_u, \vec{x}_l) : (\vec{x}_u, \vec{x}_l^*) \in \mathcal{F}_l - \mathcal{F}_{l_1} \right\}$$

where $\vec{x}_l^*$ and $\vec{x}_l$ are the solutions of current and previous generations, respectively. $\mathcal{F}_{l_1}$ is a feasible region produced by the lower level optimization function. If $\Omega(\vec{x}_u)$ is not a singleton then:

$$\Omega^{\mathrm{eo}}(\vec{x}_u) = \left\{ \vec{x}_l \in \underset{\vec{x}_l}{\operatorname{argmin}} F(\vec{x}_u, \vec{x}_l) : \vec{x}_l \in \Omega(\vec{x}_u) \right\}$$

In this case, we may not get a singleton set so we pick the "best" at random. The overall extreme optimistic model w.r.t upper level can be defined as:

min $F(\vec{x}_u, \vec{x}_l)$
Such that:
$\quad \vec{x}_l = \Omega^{\mathrm{eo}}(\vec{x}_u)$
$\quad (\vec{x}_u, \vec{x}_l) \in \mathcal{F}_u$

According to TABLE I, sample data $(\vec{x}_u, \vec{x}_{l_4})$ will be picked for $EO_F$ if previous lower level fitness is more than or equal to 112.

## 4. Bilevel Problem Optimization with Ichea

ICHEA, which is a variation of EA, is an effective and versatile constraint handling tool that has been demonstrated to perform well for benchmark static and dynamic continuous CSPs in [19, 20] and COPs in [18, 21]. ICHEA uses intermarriage-crossover operator that uses knowledge from constraints rather than blindly searching for the solution. In this particular crossover, both parents belong to different feasible regions $\mathcal{F}_i$ and $\mathcal{F}_j$ where $i \ne j$. It is also possible that a parent does not belong to any of the feasible regions $S - F$. These parents are made to come closer towards the boundary of their corresponding feasible regions to locate the overlapping regions that results in more constraints being satisfied. This iterative move can be captured as:

$$O_1 = r^i (P_2 - P_1) \tag{9}$$

where offspring $O_1$ is initially placed at position $r^1(P_2 - P_1)$ which is then iteratively moved closer to parent $P_1$ until it also satisfies the constraint(s) that $P_1$ satisfies and similarly offspring $O_2$ is designated. $r$ is a coefficient in the range $(0, 1)$ which is generally 0.5 that gives binary traversal for convergence. Exponent $i$ gets incremented from 1 to a threshold value $T$ in the sequence $\langle 1, 2, \ldots, T \rangle$. $T$ is proportional to the "vastness" of the search space which is generally $\geq 2$. The intermarriage-crossover process is shown in Fig. 2 where ✓ mark indicates possible placement for an offspring and × mark indicates the offspring vector is unacceptable in that particular position. The generated offspring from intermarriage-crossover contains genes from both parents. The purpose is to make a "generic" offspring who tries to satisfy more constraints because his parents are from two different feasible regions. The algorithm favors those offspring who satisfy more constraints by utilizing Deb's ranking scheme based on feasibility [6] to rank the individuals. The population is first sorted according to number of satisfied constraints in decreasing order then by fitness value in increasing order. The worst time complexity of this crossover is the same as the time complexity of an individual objective function evaluation $O(objFunc)$ i.e. $O(2 \times T \times objFunc) = O(objFunc)$.

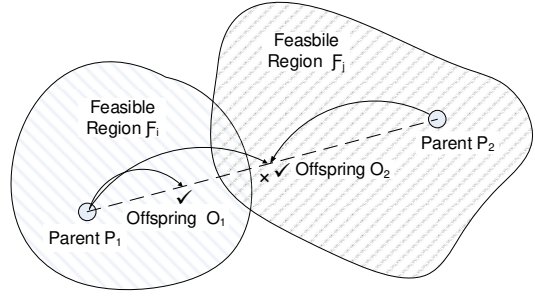

Fig. 2. Intermarriage-Crossover between parents P1 and P2

The details of complete ICHEA algorithm for bilevel problem, called Bilevel ICHEA (BICHEA) is given in Fig. 3. It has a partial nested approach where evolutionary algorithms (ICHEA) is at the upper level and local search (mutation of clones with intermarriage-crossover) at the lower level. The rest of the structure does not deviate much from the original ICHEA. BICHEA can be described in four major steps:

**Step1:** The algorithm starts with the initialization of a set of chromosomes $C$ that evolves for a given number of generations. Each chromosome contains upper and lower level input parameters, current fitness values and information about the given constraints being violated.

**Step 2:** this is a nested local search step where search for more promising lower level parameters happen. Here all feasible chromosomes go through exploitation process of hyper mutation with *localSearch* defined in Fig. 4. *localSearch* applies the optimistic convergence technique for sectors $S_2$ and $S_3$ and extreme-optimistic otherwise. We have also tried to use different orders for convergence techniques but the outcome remains same. Subroutine *clone* uses the concept of hyper-clone defined in [4]. It merely creates $\min\left(10, \; ceil\left(\frac{\beta \cdot \|C\|}{i}\right)\right)$ number of clones in proportion to the order of fitness given by $i \in \{1, \ldots, \|C\|\}$ and constant parameter $\beta$ which has been set to 0.5 for this work. Here intermarriage happens between a given chromosome with a randomly generated boundary values. For a given individual $P$ with parameters $(\vec{x}_u, \vec{x}_l)$, $\vec{x}_u$ is fixed for the lower-level to get the most suitable $\vec{x}_l$. Intermarriage-crossover creates a clone $P'$ for this individual that replaces $\vec{x}_l$ with its boundary values. Boundary value B for a variable $\vec{x}_{l_i} : 1 \leq i \leq \|\vec{x}_l\|$ is either lower bound or upper bound denoted by $B^{min}(\vec{x}_{l_i}) = min(\vec{x}_{l_i})$ and $B^{max}(\vec{x}_{l_i}) = max(\vec{x}_{l_i})$, respectively. We randomly pick the minimum or maximum for each variable in a vector $\vec{x}_l$ given by B($\vec{x}_l$). Now the distance $\delta$ between $P$ and $P'$ is:

$$\delta = \left\| \begin{bmatrix} \vec{x}_u \\ \vec{x}_l \end{bmatrix} - \begin{bmatrix} \vec{x}_u \\ B(\vec{x}_l) \end{bmatrix} \right\| = \begin{bmatrix} \mathbf{0} \\ \|\vec{x}_l - B(\vec{x}_l)\| \end{bmatrix} \tag{10}$$

$P$ move towards $P'$ in a hyper-plane to search for the local best solution. If the boundary values are not given, then any large value such as $\pm 10E4$ can be used instead.

**Step 3:** this is the upper level search with Intermarriage-crossover (*interMarCrossover*) which tries to explore for more promising chromosomes (based on a given convergence strategy) without fixing any of the parameters $(\vec{x}_u, \vec{x}_l)$ contrary to Step 2. This step is necessary to get diverse individuals in a population [21].

**Step 4:** lastly, *SortAndDiscard* filters out promising chromosomes based on given sectors with either upper level objective function $F$ or lower level objective function $f$. $S_1 - S_4$ divides the total generations in order from first to fourth sectors. Here promising solutions are selected based on a given convergence strategy which causes different convergence for each sector. Sectors $\{S_1, S_2\}$ and $\{S_3, S_4\}$ sort the population w.r.t upper level fitness, and lower level fitness respectively.

```
Function BICHEA(problem, Generations)
    C = initializeChromosomes(problem);
    For each g ∈ Generations
        C = localSearch(C,g);
        P = tournamentSelection(C);
        O = interMarCrossover(P);
        C = C ∪ O;
        sortBy = g ∈ {S₁ ∪ S₂}? byF : byf;
        C = SortAndDiscard(C, sortBy);
        PrintBest5(C);
        CheckTerminationCriteria();
    End For
End Function
```

Fig. 3. Pseudocode for BICHEA

```
Function C' = localSearch(C,g)
    C' = ∅
    For each c ∈ C
        H = clones(c);
        H' = ∅
        For (each h ∈ H)
            h' = InterMarCrossover(h,aBoundaryOf(h));
            H' = h' ∪ H'
        End for
        model = g ∈ {S₂ ∪ S₃}? Optimistic : extreme Optimistic;
        H' = sort(H', model);
        C' = C' ∪ H'.getBest();
    End for
    Return C';
End Function
```

Fig. 4. Pseudocode for subroutine *blMutation*

## 5. Experiments and Discussion

BICHEA has been tested on ten standard benchmark bilevel problems from [10, 26] to evaluate its performance with other recently developed evolutionary approaches discussed in Section 1. TABLE II describes the best known fitness values for upper level (F) and lower level (f). The problem set includes combination of linear and non-linear functions with mostly small dimensional problems with weak constraint strengths ($\rho$) apart from lower level constraint. This shows that feasible regions can be identified almost immediately. $\rho$ is computed offline by using the formula $\rho = |\text{feasible population}|/|\text{population}|$ randomly. We used a population size of 10,000 to determine the $\rho$ value as the average of five successive runs [12, 20].

TABLE II

Best Known Fitness for Benchmark Problems

| | Best known Fitness | | |
|---|---|---|---|
| Problem | F | f | $\rho$ |
| TP1 | 225.0 | 100.0 | 0.91 |
| TP2 | 0.0 | 100 | 0.50 |
| TP3 | -18.6787 | -1.0156 | 0.51 |
| TP4 | -29.2 | 3.2 | 0.11 |
| TP5 | -3.6 | -2.0 | 0.61 |
| TP6 | -1.2091 | 7.6145 | 0.04 |
| TP7 | -1.96 | 1.96 | 0.21 |
| TP8 | 0.0 | 100.0 | 0.61 |
| TP9 | 0.0 | 1.0 | 1.00 |
| TP10 | 0.0 | 1.0 | 1.00 |

TABLE III

PARAMETER SETTINGS FOR BICHEA

| Parameters | Values |
|---|---|
| Population size | 100 |
| Generations | 500 (S1: 1-125; S2: 126-250; S3: 251 – 375; S4: 376 – 500) |
| T | 2 |
| $\beta$ | 0.5 |
| blMutation rate | 1.0 |
| Crossover rate | 0.8 |
| Runs | 30/problem |
| Upper/lower bounds | $\pm 10,000.00$ (if not specified) |

TABLE IV

BEST FITNESS COMPARISON

| Prob | BICHEA | | Bayesian | | BLEAQ | | HPSOBLP | |
|---|---|---|---|---|---|---|---|---|
| | **F** | **f** | **F** | **f** | **F** | **f** | **F** | **f** |
| TP1 | **225.00009** | **99.99968** | 225.0011 | **99.9984** | 225.0 | 100.0 | 225 | 100 |
| TP2 | **0.00003** | 199.99971 | 0.0 | 200.0 | 5.4204 | 0.0 | **0** | **100** |
| TP3 | **-18.67869** | **-1.01559** | -18.6786 | -1.0156 | -18.6787 | -1.0156 | -14.8 | 0.21 |
| TP4 | **-29.19869** | 3.19633 | -29.1991 | **3.2001** | -29.2 | **3.2** | -36.0 | 0.25 |
| TP5 | **-3.67982** | -2.01346 | -3.8998 | **-2.0039** | -2.4828 | -7.705 | - | - |
| TP6 | **-1.20918** | 7.61450 | -1.2099 | 7.6173 | -1.2099 | 7.6173 | - | - |
| TP7 | **-1.96001** | **1.96001** | -1.6833 | 1.6833 | -1.8913 | 1.8913 | - | - |
| TP8 | **0.00000** | **100.00000** | 0.0 | 200.0 | 12.2529 | 0.0007 | - | - |
| TP9 | **0.00000** | **1.00000** | 0.0007 | **1.0** | 3.5373 | **1.0** | - | - |
| TP10 | **0.00000** | **1.00000** | 0.0011 | **1.0** | 0.001 | **1.0** | - | - |

TABLE V

AVERAGE FITNESS COMPARISON

| Prob | BICHEA | | Bayesian | | BLEAQ | | HPSOBLP | |
|---|---|---|---|---|---|---|---|---|
| | **F** | **f** | **F** | **f** | **F** | **f** | **F** | **f** |
| TP1 | **224.96290** | **99.99192** | 253.6155 | 70.3817 | **224.9989** | 99.9994 | 225 | - |
| TP2 | **0.00012** | 199.99989 | **0.0007** | 183.871 | 2.4352 | 93.5484 | **0** | - |
| TP3 | **-18.67862** | **-1.01533** | -18.5579 | -0.9493 | **-18.6787** | **-1.0156** | -14.0 | - |
| TP4 | **-29.15966** | 2.66120 | -27.6225 | 3.3012 | -29.2 | **3.2** | -36.0 | - |
| TP5 | -4.26669 | -1.99426 | -3.8516 | -2.2314 | **-3.4861** | -2.569 | - | - |
| TP6 | -1.21052 | **7.61467** | **-1.2097** | 7.6168 | -1.2099 | 7.6173 | - | - |
| TP7 | **-1.96173** | **1.96173** | -1.6747 | 1.6747 | -1.9538 | 1.9538 | - | - |
| TP8 | 0.36679 | **92.69204** | **0.0008** | 180.645 | 1.1463 | 132.559 | - | - |
| TP9 | **0.00000** | **1.00000** | 0.0012 | **1.0** | 1.2642 | **1.0** | - | - |
| TP10 | **0.00000** | **1.00000** | 0.0049 | **1.0** | **0.0001** | **1.0** | - | - |

TABLE VI

BEST FITNESS STATISTICS FOR BICHEA

| Prob | F | Variance for F | f | Variance for f |
|---|---|---|---|---|
| TP1 | 225.00009 | 0.033 | 99.99968 | 0.077 |
| TP2 | 0.00003 | 0.000 | 199.99971 | 0.002 |
| TP3 | -18.67869 | 0.000 | -1.01559 | 0.000 |
| TP4 | -29.19869 | 0.122 | 3.19633 | 0.174 |
| TP5 | -3.67982 | 0.621 | -2.01346 | 1.034 |
| TP6 | -1.20918 | 0.000 | 7.61450 | 0.000 |
| TP7 | -1.96001 | 0.000 | 1.96001 | 0.000 |
| TP8 | 0.00000 | 5.204 | 100.00000 | 834.887 |
| TP9 | 0.00000 | 0.000 | 1.00000 | 0.000 |
| TP10 | 0.00000 | 0.000 | 1.00000 | 0.000 |

**Best Fitness Accuracy Comparison**

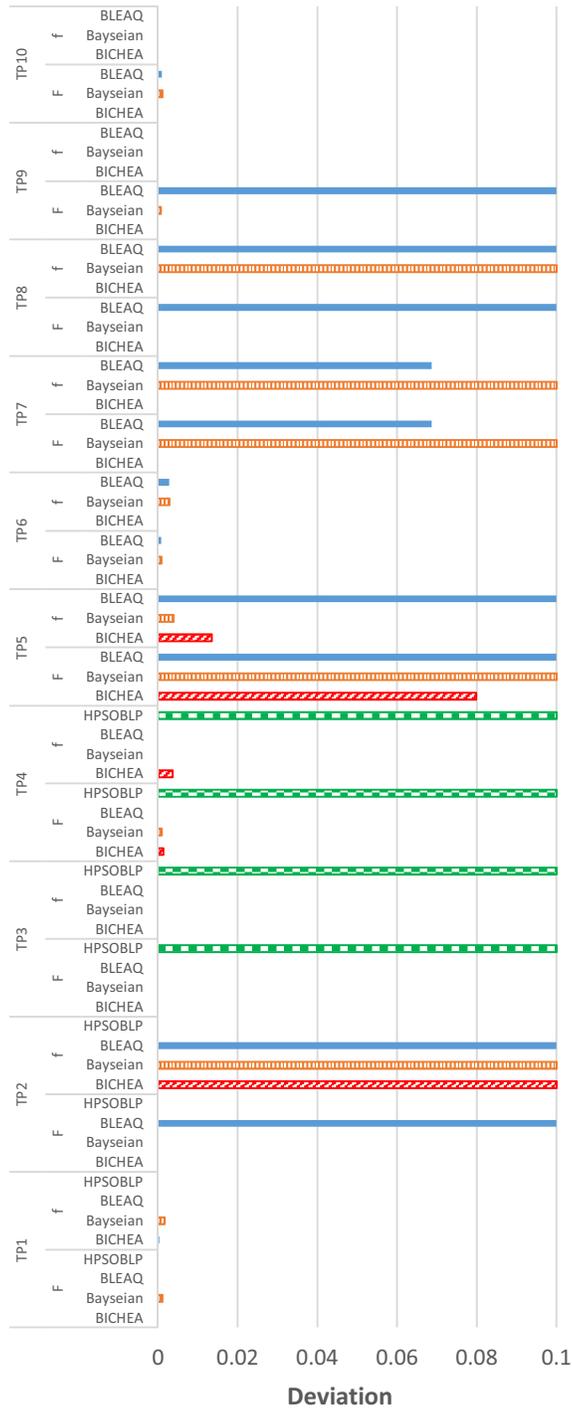

Fig. 5. Fitness accuracy of best found solutions

Parameter settings for BICHEA is shown in TABLE III where S1-S4 indicates four equal sectors with given ranges to test different convergence approaches. Sectors S1 with generations 1-125, S2 with generations 126-250, S3 with generations 251-375 and S4 with generations 276-500 have been used to test convergence approaches EO$_F$, O$_F$, O$_f$ and EO$_f$ respectively. In case, a finite range for a variable is not defined, we used upper/lower limit of value $\pm 10,000.00$. Overall, each problem was executed 30 runs consecutively to have a comparative analysis with the published results of BLEAQ, Bayesian and HPSOBLP from [10, 11]. Value of parameter $T$ can be adaptive but currently a constant value 2 has been used as it gives more promising results compared to other values. Fig. 6 shows the behavior of $T$ on the quality of solutions i.e. ($|F - F_b| + |f - f_b|$), where $\{F_b, f_b\}$ are known optimum solutions. The higher values shows degradation in the quality of the solutions which are also inefficient as the computations increases many folds for intermarriage crossover.

TABLE IV and TABLE V show the list of best and average results, respectively from all the tested algorithms. Bold values indicate the best (or almost best) among the tested algorithms and '-' indicates the unavailability of the result. BICHEA has also produced competitive results when only average fitness is considered, however, it has performed very well for best fitness values for all the testing problems except for lower level fitness in problems TP2 and TP5. For most of the problems, BICHEA has found the known best solutions. Fig. 5 shows the deviation of the best found solution $\{F, f\}$ from known optimum solutions $\{F_b, f_b\}$ i.e. $|F_b - F|$ for upper level and $|f_b - f|$ for lower level. Deviation of more than $\pm 0.1$ is considered an unacceptable solution, thus terminated to 0.1. BICHEA shows very few or smaller bars of deviation compared to the other algorithms. Overall, BICHEA performs almost equally well for most of the problems in every test runs with low variations as shown in TABLE VI. Only TP8 and somewhat TP5 do not show consistency with high variance which is also observed in TABLE V for average result.

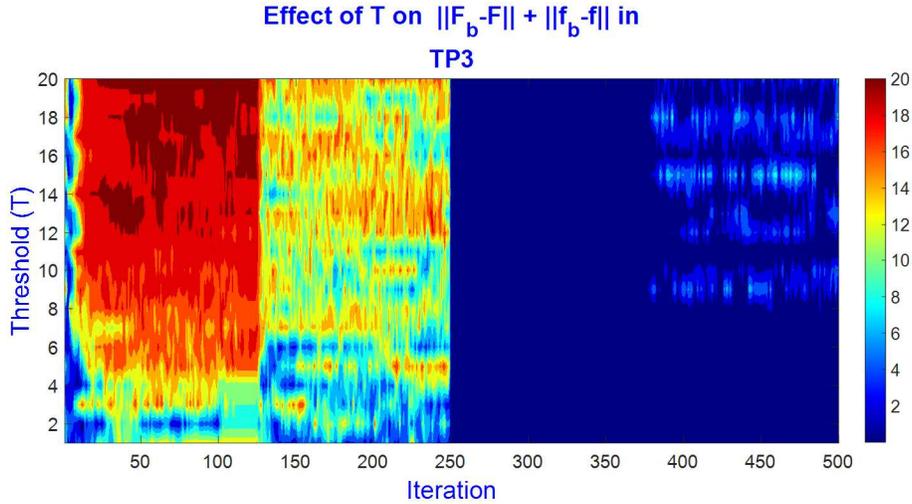

Fig. 6. Analysis on the values of threshold T w.r.t the quality of a solution

Since BICHEA runs with different convergence strategy in each sector of the total generations, we have plotted fitness landscape of the best five individuals (since the best upper level fitness value fluctuates once better lower level fitness is found) on typical runs for each of the test problems in Fig. 6 to show their impact on different convergence strategies. It can be observed that each strategy is behaving differently for the given benchmark problems to reach towards the known best solution. Optimum solutions close to best known solutions for the problems (TP1 and TP4-TP10) have been obtained in sector 2 i.e. with convergence approach of O$_F$. Notably, the convergence approaches EO$_F$ and O$_f$ have produced optimum solutions for problems {TP2, TP8-TP10} and TP3, respectively. We do not have one common convergence strategy that works best for all the problems. All but EO$_f$ have converged to known optimum solution for one or more test problems. It can also be observed that decision makers can have more than one choice of solutions as in TP1 with solution sets of (112.5, 112.5) and (225.0, 100.0) because of bimodal or multimodal nature of the problem. Similarly, (-26.2, 9.8) and (-18.7,-1.0) for TP3 and (0.0, 88.3) and (0.0, 100.0) or even (0.0, 200.0) in some cases for TP8.

We have also done non-parametric Wilcoxon test for ranking (two-tailed) with significance difference of $p < 0.05$ for all convergence strategies on every tested problem. The results for fitness values F and f are shown in TABLE VII and TABLE VIII, respectively. The results show that the difference in behavior of the convergence strategies are statistically significant in most of the cases. Especially, the prominent convergence strategies EO$_F$, O$_F$ and O$_f$ have shown the significant difference for fitness $F$. Similar results have been obtained for fitness $f$, however, problems TP9 and TP10, in particular, are not showing significant difference for varying strategies.

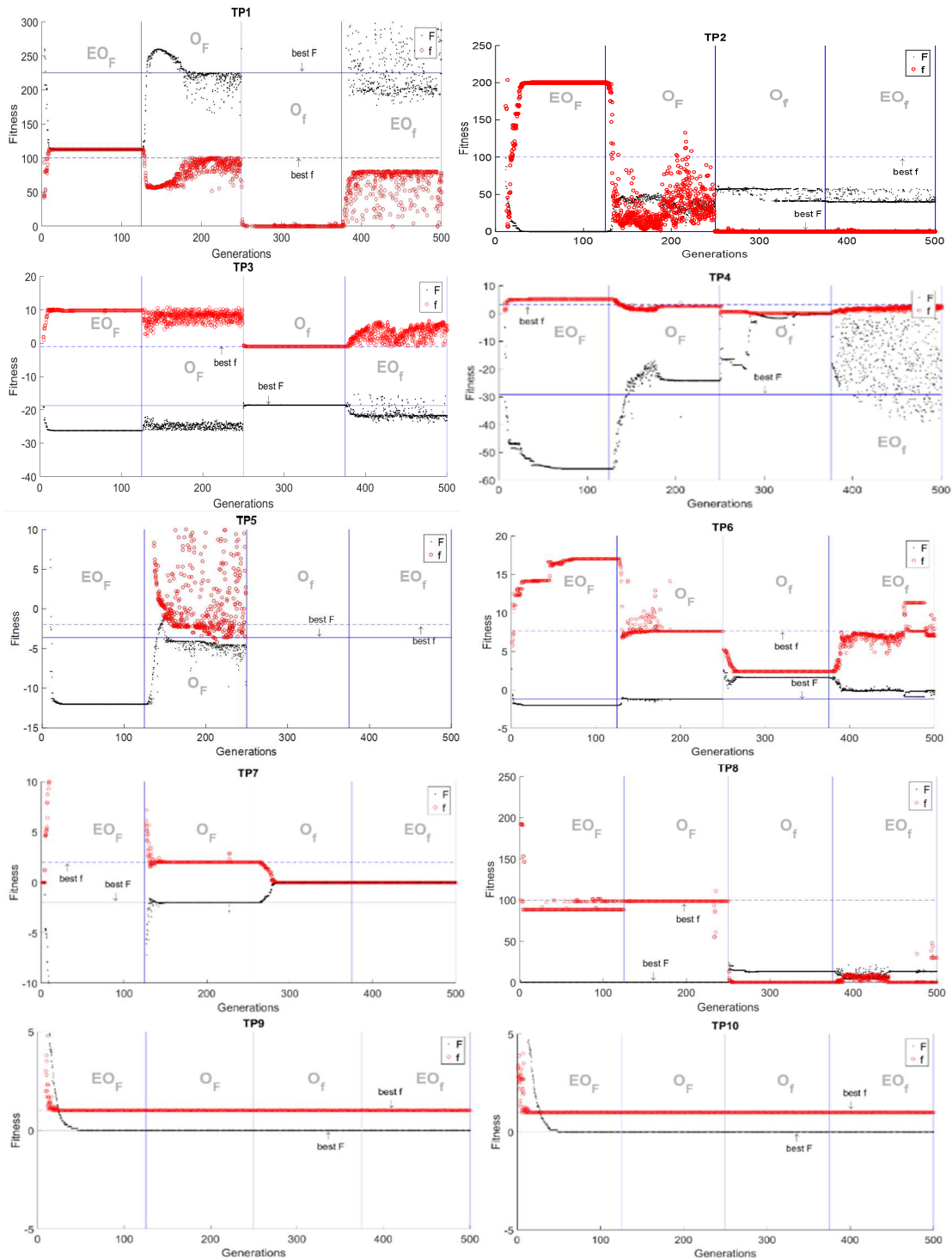

Fig. 7. Plot of best 5 fitness values w.r.t given optimistic convergence strategy in every generation

All the results discussed above are based on four variations of optimistic convergence strategies. Next, we evaluate performance of pessimistic convergence strategies and compare them with optimistic convergence strategies. TABLE X and TABLE IX compares optimistic and pessimistic approaches for the best and average fitness values obtained over 30 run on each test problem under the same conditions described in TABLE III. Bold values show better (or almost equal) result. Even though the best and average fitness values with pessimistic approaches of almost half of the test problems are matching with optimistic approaches the high variance of best fitness value on 30 runs of pessimistic approach is a concern. Fig. 8 shows the difference of variance $\Delta var(F) = var(F_O) - var(F_P)$ and $\Delta var(f) = var(f_O) - var(f_P)$ where subscripts $O$ and $P$ refer to optimistic and pessimistic approach respectively. The positive value indicates that the variance of optimistic approach is high and vice versa for negative value. Pessimistic approach has higher variance for all the problem except for TP8 where it has produced a very high variance. Difference of variance of more than $\pm 1$ is terminated to $\pm 1$. TP9 and TP10 have a variance value of 0.0.

TABLE VII

CROSS EVALUATION MATRIX OF CONVERGENCE STRATEGIES FOR FITNESS $F$ BASED ON WILCOXON RANK TEST ($p < 0.05$)

| Conv. | $EO_F$ | $O_F$ | $O_f$ | $EO_f$ |
|---|---|---|---|---|
| $EO_F$ | - | TP1-TP7, TP9, TP10 | TP1-TP10 | TP1-TP10 |
| $O_F$ | TP1-TP7, TP9, TP10 | - | TP1-TP8 | TP2-TP8 |
| $O_f$ | TP1-TP10 | TP1-TP8 | - | TP1,TP2, TP4,TP6,TP8 |
| $EO_f$ | TP1-TP10 | TP2-TP8 | TP1,TP2, TP4, TP6, TP8 | - |

TABLE VIII

CROSS EVALUATION MATRIX OF CONVERGENCE STRATEGIES FOR FITNESS $f$ BASED ON WILCOXON RANK TEST ($p < 0.05$)

| Conv. | $EO_F$ | $O_F$ | $O_f$ | $EO_f$ |
|---|---|---|---|---|
| $EO_F$ | - | TP1-TP7 | TP1-TP8 | TP1-TP8 |
| $O_F$ | TP1-TP7 | - | TP1-TP8 | TP1-TP8 |
| $O_f$ | TP1-TP8 | TP1-TP8 | - | TP1-TP4, TP6, TP8 |
| $EO_f$ | TP1-TP8 | TP1-TP8 | TP1-TP4, TP6, TP8 | - |

TABLE IX

BEST FITNESS COMPARISON OF OPTIMISTIC VS PESSIMISTIC APPROACHES

| | Pessimistic | | Optimistic | |
|---|---|---|---|---|
| Prob | F | f | F | f |
| TP1 | 223.62584 | 99.64207 | **225.00009** | **99.99968** |
| TP2 | **0.00002** | 199.99983 | 0.00003 | 199.99971 |
| TP3 | **-18.6787** | **-1.01559** | -18.67869 | -1.01559 |
| TP4 | -29.27955 | 3.55248 | **-29.19869** | **3.19633** |
| TP5 | -4.28568 | -0.32331 | **-3.67982** | **-2.01346** |
| TP6 | -1.21742 | 7.62715 | **-1.20918** | **7.61450** |
| TP7 | -1.9612 | 1.9612 | **-1.96001** | **1.96001** |
| TP8 | **0.00000** | **100.00000** | 0.00000 | 100.00000 |
| TP9 | **0.00000** | **1.00000** | 0.00000 | 1.00000 |
| TP10 | **0.00000** | **1.00000** | 0.00000 | 1.00000 |

TABLE X

AVERAGE FITNESS COMPARISON OF OPTIMISTIC VS PESSIMISTIC APPROACHES

| | Pessimistic | | Optimistic | |
|---|---|---|---|---|
| Prob | F | f | F | f |
| TP1 | 205.06778 | 87.70457 | **224.96290** | **99.99192** |
| TP2 | **0.00012** | **199.99894** | 0.00012 | 199.9989 |
| TP3 | **-18.67884** | **-1.01559** | -18.67862 | -1.01533 |
| TP4 | -29.06566 | 4.27437 | **-29.15966** | 2.66120 |
| TP5 | **-3.12431** | 1.80181 | -4.26669 | **-1.99426** |
| TP6 | -1.11267 | 7.6301 | **-1.21052** | **7.61467** |
| TP7 | -1.90907 | 1.90907 | **-1.96173** | **1.96173** |
| TP8 | 0.3518 | 106.28977 | 0.36679 | 92.69204 |
| TP9 | **0.0000** | **1.0000** | 0.00000 | 1.00000 |
| TP10 | **0.0000** | **1.0000** | 0.00000 | 1.00000 |

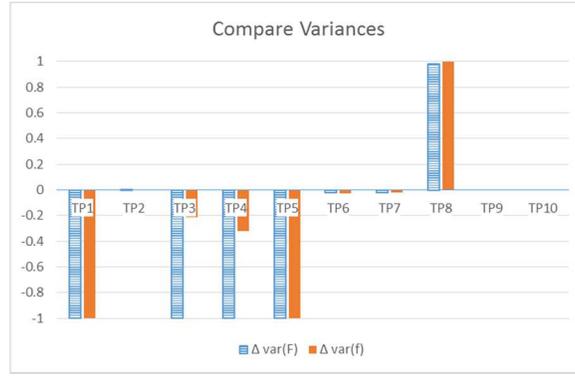

Fig. 8. Comparison of difference of variance for optimistic and pessimistic approaches.

Additionally, the statistical significance of the difference between these two approaches on BICHEA is tabulated in TABLE XI and TABLE XII for fitness $F$ and $f$, respectively, with two-tailed Wilcoxon rank test with significance difference of $p < 0.05$. Gens indicate the generation number on which the fitness values are collected for statistical analysis. Major difference is observed between $P_F$ and $O_F$ where majority of the best fitness values have been obtained as described earlier, however, $EP_F$ and $EO_F$, and $P_f$ and $O_f$ did not show any major difference for most of the problems. $EP_f$ and $EO_f$ show difference in convergence but both are weak convergence techniques as far as attaining close to optimum solution is concerned. We have plotted average fitness of both levels vs generation graph for convergence strategies $P_F$ and $O_F$ for some of the problems in Fig. 9 and Fig. 10 to show how these most prominent convergence strategies behave differently on the same problem. The graph of the selected problems TP1 and TP6 shows that $P_F$ is generally stuck in local best for most of the generations because of its greedy behavior of selecting the best $f$ value that gives worst $F$ solution. On the other hand, $O_F$'s greediness towards better $F$ solution leads it towards the known best solution. For TP1, major difference in fitness $F$ can be observed while in TP6, both strategies eventually converge close to known best, however, they differ a lot in their convergence approaches. $O_F$ can also be observed converging faster than $P_F$.

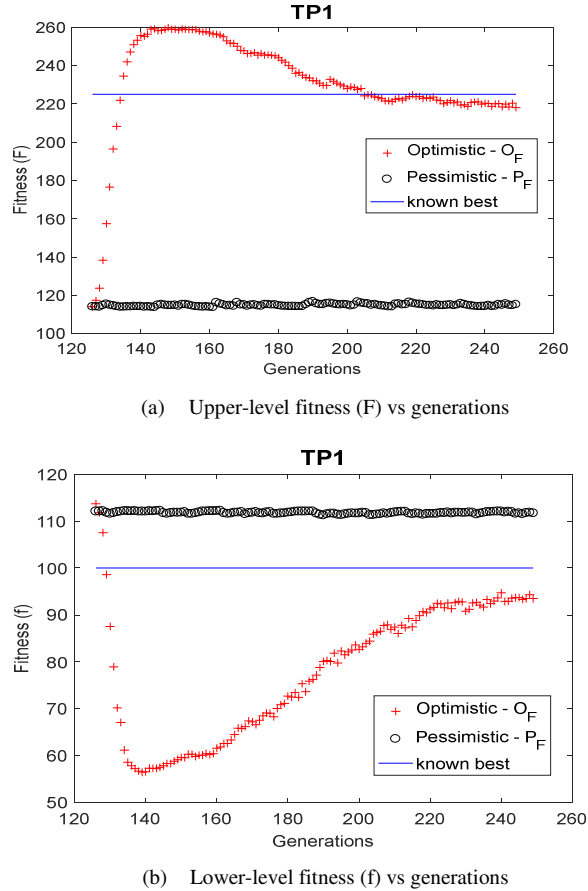

(a)    Upper-level fitness (F) vs generations

(b)    Lower-level fitness (f) vs generations

Fig. 9. Plot of average fitness values for problem TP1 w.r.t. optimistic strategy $O_F$ and pessimistic strategy $P_F$

Finally, the efficiency of BICHEA can be formulated with the average time complexity of blMutation for $T = 20$. It evaluates to $O(N(20 \times objFunc + 20log20)) = O(N \times objFunc)$ where N is population size and average sorting time complexity of $n$ sized problem is taken as $nlogn$. Thus the time complexity of overall BICHEA is $O(G(2NobjFunc + NlogN + 2N)) = O(NG(objFunc + logN))$ where G is the total generation, which can be inversely proportional to $N$ in many general cases for EAs. In that case, the time complexity would be simply $O(objFunc + logN)$ or $O(objFunc)$ when tested with constant population size $N$.

TABLE XI

COMPARISON OF OPTIMISTIC VS PESSIMISTIC APPROACHES FOR
FITNESS $F$ WITH WILCOXON RANK TEST

| Conv. | Gens | Wilcoxon rank test ($p < 0.05$) |
|---|---|---|
| $EP_F \mid EO_F$ | 125 | TP4, TP7 |
| $P_F \mid O_F$ | 250 | TP1 - TP7 |
| $P_f \mid O_f$ | 375 | TP7, TP8 |
| $EP_F \mid EO_f$ | 500 | TP1, TP2, TP4 - TP6, TP8 |

TABLE XII

COMPARISON OF OPTIMISTIC VS PESSIMISTIC APPROACHES FOR
FITNESS $f$ WITH WILCOXON RANK TEST

| Conv. | Gens | Wilcoxon rank test ($p < 0.05$) |
|---|---|---|
| $EP_F \mid EO_F$ | 125 | TP4, TP7 |
| $P_F \mid O_F$ | 250 | TP1 - TP7 |
| $P_f \mid O_f$ | 375 | TP2, TP4, TP7 |
| $EP_F \mid EO_f$ | 500 | TP1,TP3, TP4, TP6 |

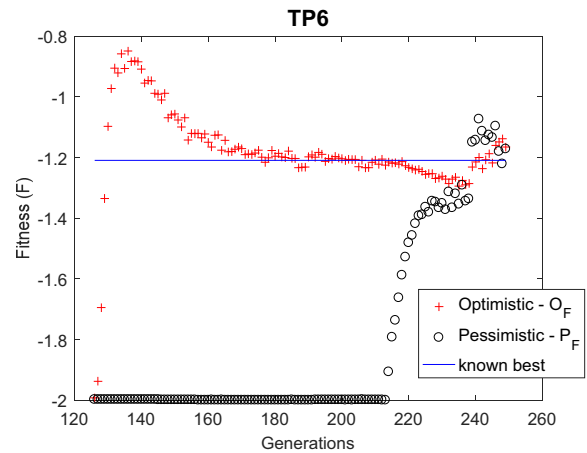

(a)   Upper-level fitness (F) vs generations

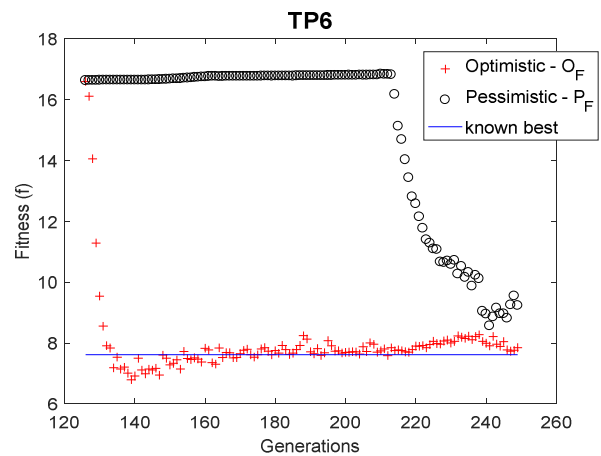

(b)   Lower-level fitness (F) vs generations

Fig. 10. Plot of average fitness values for problem TP6 w.r.t. optimistic strategy $O_F$ and pessimistic strategy $P_F$

## 6. Conclusion and Future Work

Bilevel problem is a class of constraint optimization problem where one of the constraints is an optimization function. Earlier mathematical programming was used to solve these problems, but recently, few partial and complete evolutionary computation approaches have been proposed. Our proposed algorithm BICHEA is a complete evolutionary approach with a single level optimization structure with intermarriage-crossover. It was compared with BLEAQ and Bayesian as partial evolutionary approaches and HPSOBLP as a complete evolutionary approach. BICHEA has outperformed other algorithms in terms of quality of solutions. BICHEA was able to reach towards known global (or near global) optimum solution for all the tested benchmark problems. In this paper, we have realized that different forms of convergence approaches behave differently; however, we have only tested the optimistic approach and its variants. It was demonstrated that our proposed optimistic variants, namely $EO_F$ and $O_F$, have produced global optimal solutions with BICHEA for some of the problems where $O_F$ was unable to do so. Future work would be considered for BICHEA for multi-objective bilevel optimization, which is even a more complex to deal with, as both levels can have multiple objectives to solve. Here, feasibility of any given upper level variable is determined by producing lower level pareto front. Upper level needs to produce the optimal pareto front for the final solution. Some other forms of bilevel problem can be dynamic lower level constraints and discrete search space.

## ACKNOWLEDGMENTS

I would like to thank Prof. Grégoire Danoy and Emmanuel Kieffer for providing the data files from their experimental results from [10].